\documentclass[aps,pre,twocolumn,superscriptaddress,showpacs]{revtex4-2}
\usepackage{graphicx}
\usepackage{array}
\usepackage{amsfonts,amssymb,amsmath,mathrsfs,multirow,rotate}
\usepackage{booktabs}
\usepackage{soul}
\usepackage{threeparttable}
\usepackage{multirow}
\usepackage{epsfig}
\usepackage{threeparttable}
\usepackage{chngpage}
\usepackage{latexsym}
\usepackage{dcolumn}
\usepackage{graphics,graphicx,bm,fleqn,epic,eepic,float}
\usepackage{verbatim}
\usepackage{color}
\usepackage{xr}
\usepackage{float}
\usepackage{listings} 
\usepackage{placeins} 
\usepackage{tikz}
\usepackage{url}
\usepackage{natbib}

\begin{document}

\title{Data-driven model discovery with Kolmogorov-Arnold networks}

\date{\today}

\author{Mohammadamin Moradi} \email{The first two authors contributed equally to this work}
\affiliation{School of Electrical, Computer, and Energy Engineering, Arizona State University, Tempe, AZ 85287, USA}

\author{Shirin Panahi} \email{The first two authors contributed equally to this work}
\affiliation{School of Electrical, Computer, and Energy Engineering, Arizona State University, Tempe, AZ 85287, USA}

\author{Erik M. Bollt}
\affiliation{Department of Electrical and Computer Engineering, Clarkson Center for Complex Systems Science, Clarkson University, Potsdam, New York 13699, USA}

\author{Ying-Cheng Lai} \email{Ying-Cheng.Lai@asu.edu}
\affiliation{School of Electrical, Computer, and Energy Engineering, Arizona State University, Tempe, AZ 85287, USA}
\affiliation{Department of Physics, Arizona State University, Tempe, Arizona 85287, USA}

\begin{abstract}

Data-driven model discovery of complex dynamical systems is typically done using sparse optimization, but it has a fundamental limitation: sparsity in that the underlying governing equations of the system contain only a small number of elementary mathematical terms. Examples where sparse optimization fails abound, such as the classic Ikeda or optical-cavity map in nonlinear dynamics and a large variety of ecosystems. Exploiting the recently articulated Kolmogorov-Arnold networks, we develop a general model-discovery framework for any dynamical systems including those that do not satisfy the sparsity condition. In particular, we demonstrate non-uniqueness in that a large number of approximate models of the system can be found which generate the same invariant set with the correct statistics such as the Lyapunov exponents and Kullback–Leibler divergence. An analogy to shadowing of numerical trajectories in chaotic systems is pointed out.

\end{abstract}

\maketitle

Discovering the model of a system from observational or measurement data has been
a fundamental problem since the beginning of science. For nonlinear dynamical 
systems, data-driven identification, and forecasting have attracted a great deal of 
research in the past four decades~\cite{FS:1987,CM:1987,Casdagli:1989,SGMCPW:1990,KR:1990,GS:1990,Gouesbet:1991,TE:1992,BBBB:1992,Longtin:1993,Murray:1993,Sauer:1994,Sugihara:1994,FK:1995,Parlitz:1996,SSCBS:1996,Szpiro:1997,HKS:1999,Bollt:2000,HKMS:2000,Sello:2001,MNSSH:2001,Smith:2002,Judd:2003,Sauer:2004,YB:2007,TZJ:2007,WYLKG:2011,WLGY:2011,WYLKH:2011,SNWL:2012,SWL:2012,SWL:2014,SLWD:2014,SWFDL:2014,SWWL:2016,ASB:2020}. A diverse array of methodologies have been developed, e.g., calculating the 
information contained in sequential observations to deduce the deterministic 
equations~\cite{CM:1987}, approximating a nonlinear system by a large collection of 
linear equations~\cite{FS:1987,Gouesbet:1991,Sauer:1994}, fitting differential 
equations to chaotic data~\cite{BBBB:1992}, exploiting chaotic 
synchronization~\cite{Parlitz:1996} or genetic algorithms~\cite{Szpiro:1997,TZJ:2007},
inverse Frobenius-Perron approach to designing a dynamical system ``near'' the 
original system~\cite{Bollt:2000}, or using the least-squares best approximation
for modeling~\cite{YB:2007}. An approach that has gained considerable interest 
is sparse optimization, where the system functions are assumed to have a sparse 
structure in that they can be represented by a small number of elementary mathematical 
functions, e.g., a few power- and/or Fourier-series terms. What is needed then is to 
estimate the coefficients associated with these terms. In a high-order series expansion, 
the coefficients with the vast majority of the terms are zero, except for a few. The 
problem of finding these nontrivial coefficients can then be naturally 
formulated~\cite{WYLKG:2011,YLG:2012} as a compressive-sensing 
problem~\cite{CRT:2006a,CRT:2006b,Donoho:2006,Baraniuk:2007,CW:2008}. Under the same
idea, a popular method was later developed~\cite{BPK:2016,Lai:2021}. 

The sparse-optimization approach is effective for systems whose governing equations
are sufficiently simple in the sense of sparsity, such as the chaotic 
Lorenz~\cite{Lorenz:1963} and R\"{o}ssler~\cite{Rossler:1976} oscillators whose 
velocity fields contain a small number of low-order power-series terms. However, 
sparsity can be self-sabotage because, while it is the reason that the approach is 
powerful, it also presents a fundamental limitation: it works only if the system 
equations do in fact have a sparse structure. Dynamical systems violating the sparsity 
condition arise in physical and biological situations. A known example is the Ikeda map 
that describes the propagation of an optical pulse in a cavity with a nonlinear 
medium~\cite{Ikeda:1979,HJM:1985}, whose functions contain an infinite number of series 
expansion terms. Many ecological systems and gene-regulatory circuits whose governing 
equations have a Holling-type of structure~\cite{Holling:1959a,Holling:1959b} also 
violate the sparsity condition~\cite{JHSLGHL:2018}. For these systems, the 
sparse-optimization approach to model discovery fails absolutely and completely. 

\begin{figure} [ht!]
\centering
\includegraphics[width=0.9\linewidth]{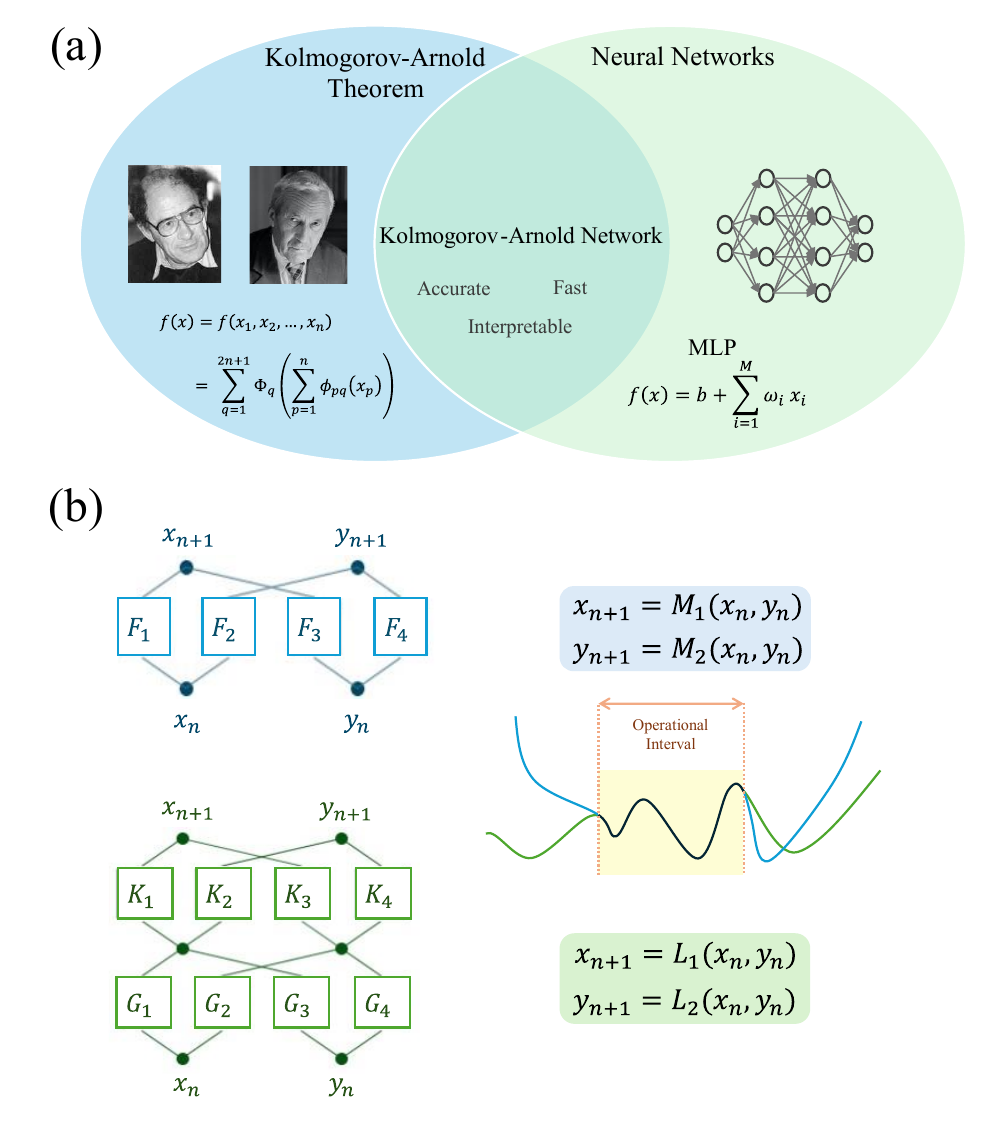}
\caption{Basics of KAN. (a) Kolmogorov-Arnold theorem and neural network. (b) Schematic 
illustration of two different structures (blue and green) leading to two different 
functions $\mathbf{M}(\mathbf{x})$ and $\mathbf{L}(\mathbf{x})$ that generate the same 
dynamics as $\mathbf{x}_{n+1} = \mathbf{F}(\mathbf{x}_n)$ in the relevant phase-space 
domain.}
\label{fig:KAN}
\end{figure}

In this Letter, we articulate an entirely different approach to discovering the models of 
any dynamical systems including those that do not meet the sparsity condition. The idea 
exploits Kolmogorov-Arnold networks (KANs), a recent computational framework for 
representing sophisticated mathematical functions~\cite{LWVRHSHT:2024} based on the 
classical Kolmogorov-Arnold theorem~\cite{Kolmogorov:1957,Arnold:2009,BG:2009} 
that any multivariate mathematical function can be decomposed as a sum of 
single-variate functions, as illustrated in Fig.~\ref{fig:KAN}(a). KANs decompose
complex high-dimensional problems into simpler, more manageable univariate functions,
allowing for more efficient training and better interpretability of
the machine-learning model, addressing some of the limitations in traditional neural
networks such as the black-box nature and computational inefficiency. As a result, KANs
are rapidly gaining attention as a promising alternative in machine learning. 

In contrast to a standard neural network with thousands and perhaps millions of 
weights and biases but always fixed all the same activation functions say $atan$ or 
$ReLu$, a KAN is a small network of say a dozen nodes but each different and carefully 
designed activation functions. We consider a dynamical system described by 
$d\mathbf{x}/dt = \mathbf{F}(\mathbf{x})$ or alternatively by 
$\mathbf{x}_{n+1}=\mathbf{F}(\mathbf{x}_n)$, including where 
$\mathbf{F}(\mathbf{x})$ {\em does not} possess a sparse structure. Our goal is to find 
an approximation of $\mathbf{F}(\mathbf{x})$, denoted as $\mathbf{G}(\mathbf{x})$, such 
that the system 
$d\mathbf{x}/dt = \mathbf{G}(\mathbf{x})$ or $\mathbf{x}_{n+1}=\mathbf{G}(\mathbf{x}_n)$
produces the identical dynamical behaviors as the original system (e.g., the same 
attractor with the same statistical and dynamical invariants to within certain numerical 
precision). We demonstrate, using the Ikeda map and a chaotic ecosystem as illustrative 
examples, that such a function $\mathbf{G}(\mathbf{x})$ in an implicit form can indeed 
be found by the KANs.

The interpretability of the KAN structure lies in its accessibility to the internal 
mechanisms of the model, such as the activation functions $F_i(\mathbf{x})$, 
$G_i(\mathbf{x})$, and $K_i(\mathbf{x})$ in Fig.~\ref{fig:KAN}(b). 
Unlike the conventional machine-learning methods, which often operate as ``black boxes,'' 
KANs provide a more transparent view of how inputs are transformed into outputs. This 
transparency allows for a better understanding of the underlying dynamics and how each 
function influences the system's behavior, thereby making them more interpretable than 
conventional methods. Figure~\ref{fig:KAN}(b) presents two different KAN structures 
highlighted in blue and green. The blue KAN has two inputs and two outputs without 
any hidden nodes, where the functions $M_1 = F_1(\mathbf{x_n}) + F_3(\mathbf{y_n})$ 
and $M_2 = F_2(\mathbf{x_n}) + F_4(\mathbf{y_n})$ are linear combinations of the 
activation functions $F_i$ for $i =1,\cdots,4$. The green structure has two extra 
hidden nodes where 
$L_1 = K_1 \big( G_1(\mathbf{x_n}) + G_3(\mathbf{y_n}) \big) + K_3 \big(G_2(\mathbf{x_n}) + G_4(\mathbf{y_n}) \big)$ 
and $L_2 = K_2 \big(G_1(\mathbf{x_n}) + G_3(\mathbf{y_n})\big)+K_4 \big(G_2(\mathbf{x_n})+G_4(\mathbf{y_n})\big)$. 
Both structures produce the same dynamics in the relevant phase-space domain (yellow 
shaded area), where the dynamics outside of this domain can be different. This concept 
will be elucidated below with a concrete example.

From the standpoint of data-driven model discovery, the Ikeda map represents perhaps 
the most difficult kind of system - so far there has been no success with any sparse 
optimization method. The two-dimensional map is given by~\cite{Ikeda:1979,HJM:1985}
$x_{n+1} = 1 + \mu \left( x_n \cos(\phi_n) - y_n \sin(\phi_n) \right)$ and
$y_{n+1} = \mu \left( x_n \sin(\phi_n) + y_n \cos(\phi_n) \right)$, where 
$\phi_n = 0.4 - 6(1 + x_n^2 + y_n^2)^{-1}$ and $\mu$ is a bifurcation parameter.
(We fix $\mu = 0.9$, so that the map generates a chaotic attractor in the phase-space 
domain $(x \in [-1,2], y\in [-2.5, 1])$. Sparse optimization fails spectacularly for 
this system because in either the power- or the Fourier-series expansions or a 
combination of both, an infinite number of terms are required to represent each map 
function - see Supplementary Information (SI) for more details~\cite{SI}.

\begin{figure} [ht!]
\centering
\includegraphics[width=0.9\linewidth]{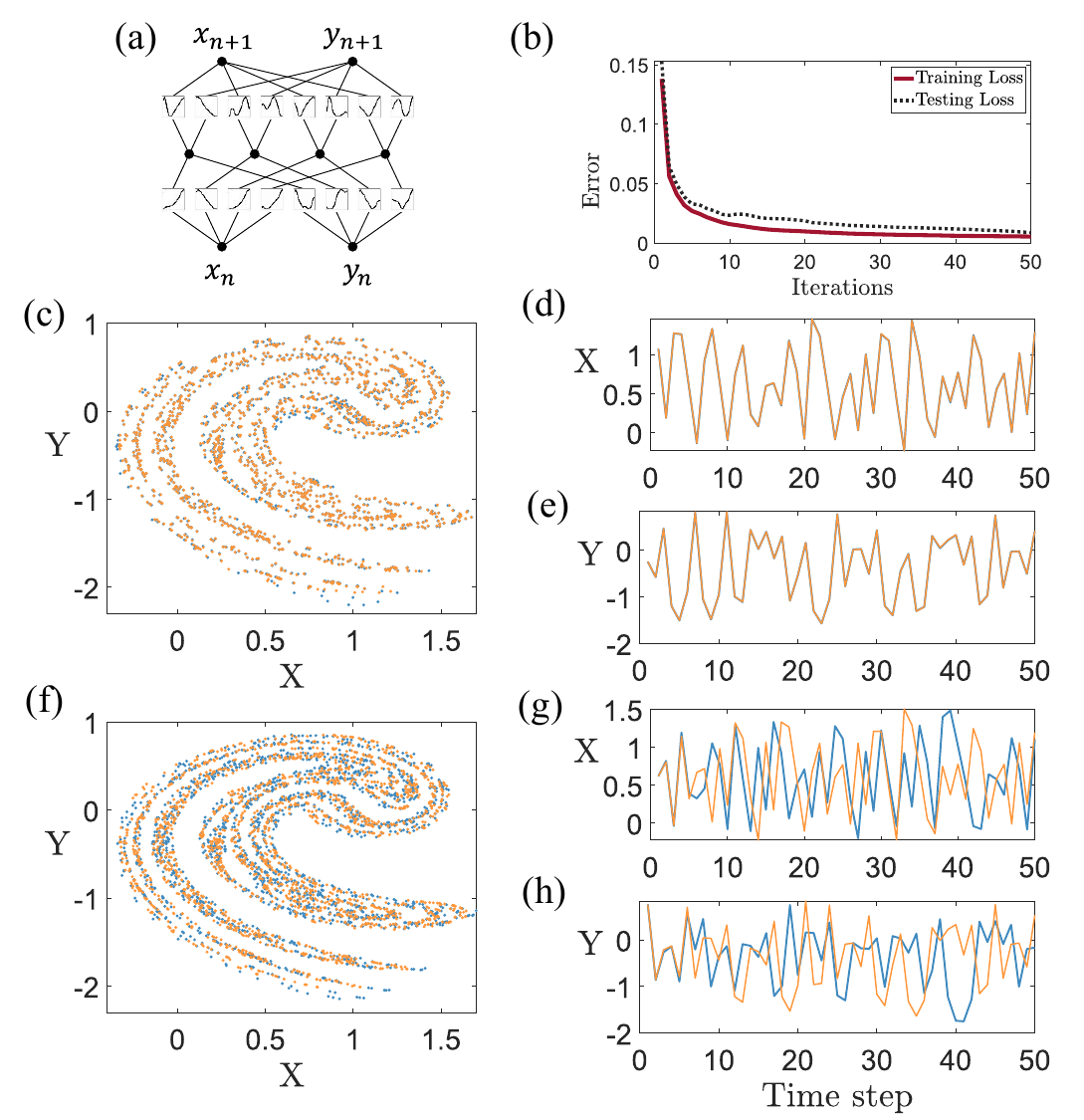}
\caption{KANs applied to the Ikeda map. (a) A KAN structure with 2 input, 4 hidden, 
and 2 output nodes. (b) Training (red) and testing (black dashed) loss curves. 
(c) Chaotic attractor during the training phase (blue - ground truth; orange - KAN 
produced). (d,e) Time series during the training. The blue and orange traces overlap 
well, signifying a high training accuracy. (f) Chaotic attractor during testing (blue - 
ground truth; orange - KAN produced). (g,h) The corresponding time series. While the 
predicted time series diverges from the ground truth after a few iterations due to chaos, 
the KAN generates the correct attractor in the pertinent phase-space domain.
The true Lyapunov exponents of the chaotic attractor are $[0.5025,-0.7263]$.
The KAN predicted model gives the values of the two exponents as $[0.5075,-0.7182]$,
agreeing with the ground truth.}
\label{fig:Ikeda1}
\end{figure}

We first use a $[2,4,2]$ KAN structure, as shown in Fig.~\ref{fig:Ikeda1}(a), which
has $2$ input, $4$ hidden, and $2$ output nodes. The time-series data contain $10^4$ 
points, with 80\% allocated for training and the remaining 20\% for testing. The 
training process contains 50 iterations with the following hyperparameter values: $k=3$ 
(cubic B-splines), grid size $G=10$ for the splines, regularization parameters $\lambda=0$ 
and $\lambda_{\rm entropy} = 10$, learning rate 0.1, and a zero initial random seed.
(see SI~\cite{SI} for a detailed description of these hyperparameters). Training is 
administered in a feedforward process in which the KAN is trained to minimize the 
difference between the input and output so as to predict the evolution of the Ikeda 
map into the future with the input of the dynamical variables from the past. The training 
loss as a function of time is shown as the red curve in Fig.~\ref{fig:Ikeda1}(b), and the 
KAN-produced attractor and time series during the training phase in comparison with the 
ground truth are shown in Figs.~\ref{fig:Ikeda1}(c-e), respectively. The training loss 
decreases rapidly to zero, indicating high training accuracy and efficiency with skill. 
For the testing phase, we use the same set of parameter values but replace the original 
input data point with the output of the KAN at each iteration. The testing loss is shown 
in Fig.~\ref{fig:Ikeda1}(b) as the black dashed curve, and the KAN predicted attractor 
and time series are shown in Figs.~\ref{fig:Ikeda1}(f-h), respectively. While the  
KAN-predicted time series diverges from the ground truth after a few iterations 
due to chaos, the predicted attractor agrees with the ground truth well, indicating that 
the KAN has generated the correct model of the Ikeda map.

\begin{figure} [ht!]
\centering
\includegraphics[width=0.9\linewidth]{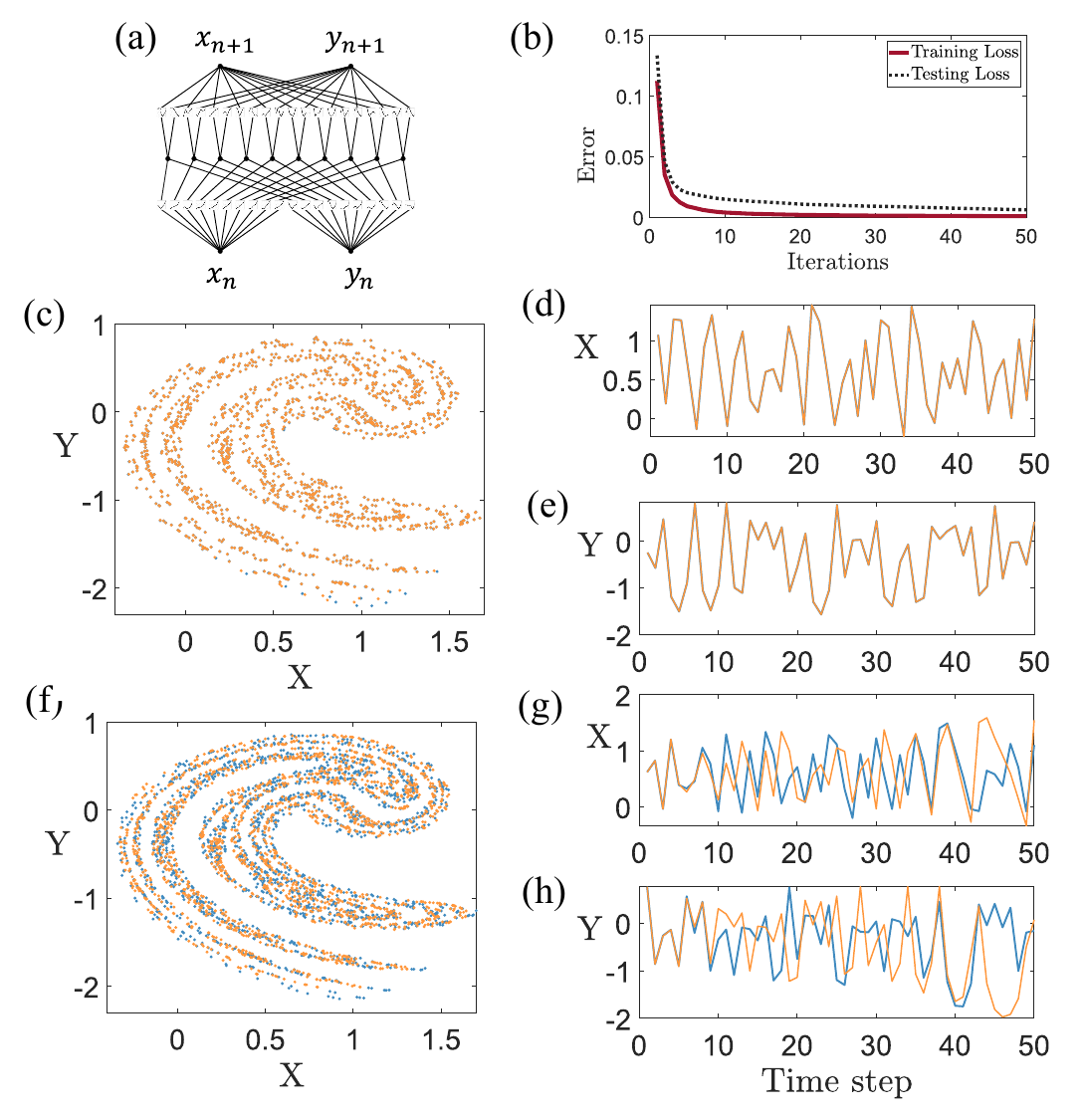}
\caption{A KAN configuration generating a different representation of the Ikeda map
but with the same chaotic attractor. The KAN has 2 input, 10 hidden, and 2 output
nodes. Legends are the same as those in Fig.~\ref{fig:Ikeda1}. The two Lyapunov 
exponents of the KAN predicted model are $[0.5033,-0.7311]$, which again agrees with
the true exponents.}
\label{fig:Ikeda2}
\end{figure}

To demonstrate that a KAN can be readily modified to generate a different representation 
of the Ikeda map but with the same chaotic attractor, we construct a more sophisticated
architecture than the one in Fig.~\ref{fig:Ikeda1}(a), as shown in Fig.~\ref{fig:Ikeda2}(a). The training and prediction results are shown in Figs.~\ref{fig:Ikeda2}(b-h).

For generality, we now present results from a continuous-time system, a chaotic 
ecosystem~\cite{MCcann1994} of three dynamical variables: 
$\dot{N} = N \left(1 - N/K\right) - x_p y_p NP/(N + N_0)$, 
$\dot{P} = x_p P \left( y_p N/(N + N_0) -1 \right) - x_q y_q PQ/(P + P_0)$, 
and $\dot{Q} = x_q Q \left( y_q P/(P + P_0) -1 \right)$,
where $N$, $P$, and $Q$ are the populations of the primary producer, the herbivore, 
and the carnivore, respectively, and the bifurcation parameter $K$ is the carrying 
capacity. For $K = 0.98$ and other parameters set as $x_p=0.4$, $y_p=2.009$, $x_q=0.08$, 
$y_q=2.876$, $N_0=0.16129$, and $P_0=0.5$, the system exhibits a chaotic 
attractor~\cite{MCcann1994}. A power-series expansion of the velocity field contains 
an infinite number of terms, violating the sparsity condition - see SI for more 
details~\cite{SI}.

\begin{figure} [ht!]
\centering
\includegraphics[width=0.9\linewidth]{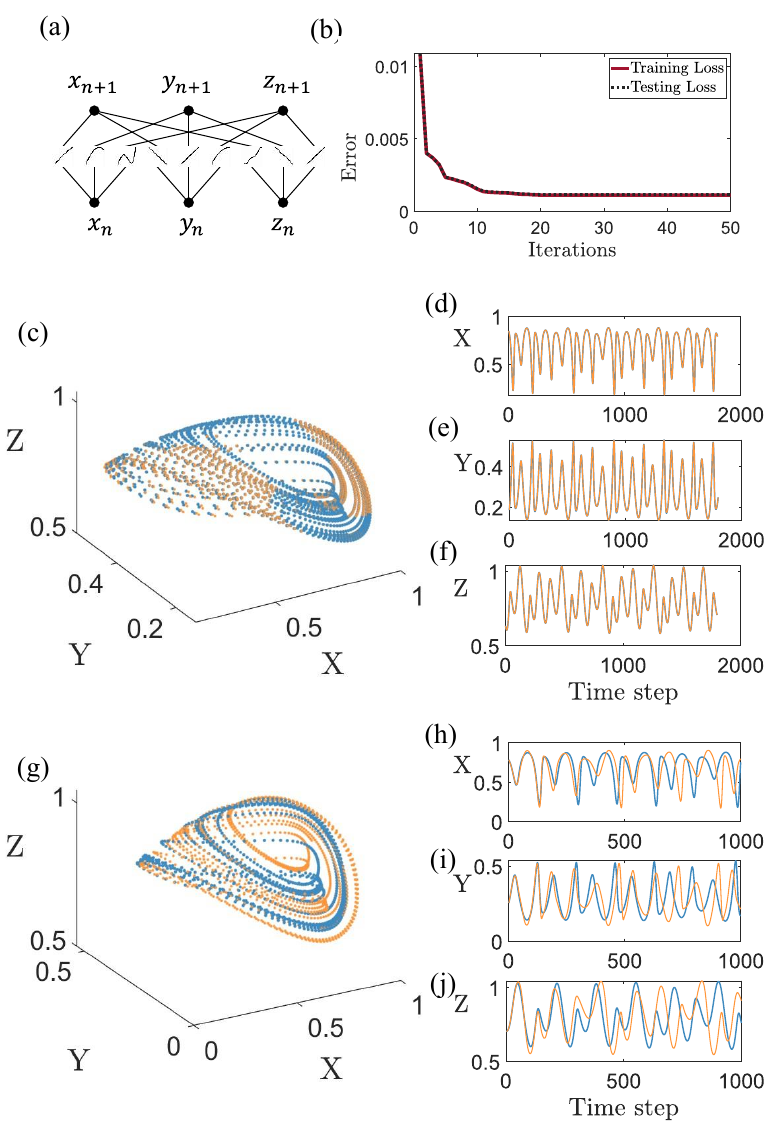}
\caption{KAN applied to a chaotic ecosystem. (a) KAN structure with 3 input and 3 
output nodes. (b) Training and testing loss curves. (c) KAN generated attractor 
during the training phase (orange), which agrees completely with the ground truth 
(blue). (d-f) KAN generated time series (orange) in agreement with the true 
time series (blue). (g-j) Similar to (c-f) but for the testing phase. Due to chaos, 
the KAN generated time series diverges from the true ones from the same initial 
condition, but the KAN attractor agrees with the true one. The true Lyapunov exponents 
are $[0.0053, 0, -0.2288]$. The exponents of the KAN-generated attractor are
consistent: $[0.0095, -5.8\times 10^{-6}, -0.3932]$. The errors arise from the 
implicit numerical evaluation of the Jacobian matrix.}
\label{fig:Food_chain}
\end{figure}

Our KAN architecture has a $[3,3]$ structure (3 input and 3 output nodes, no hidden
nodes), as illustrated in Fig.~\ref{fig:Food_chain}(a). The neural network was trained 
using 10,000 data points of sampling interval $\delta t = 0.5$ 
(corresponding to about 1,155 cycles of oscillation), with 90\% of the data allocated 
for training and the remaining 10\% for testing. The training process involved 100 
iterations for the following hyperparameter values: cubic B-spline ($K=3$), grid size 
$G=3$, $\lambda = 0$, $\lambda_{\rm entropy} = 10$, learning rate 0.5, and a zero 
initial random seed. Figure~\ref{fig:Food_chain}(b) shows the rapid decrease in the 
training and testing loss with increasing epochs. The KAN generated attractor and the 
corresponding time series during the training phase are shown in 
Figs.~\ref{fig:Food_chain}(c-f), where a comparison with the ground truth indicates 
successful training. The KAN attractor and the time series generated during the testing 
phase are shown in Figs.~\ref{fig:Food_chain}(g-j), demonstrating the KAN's forecasting 
power. The Lyapunov exponents of the attractor are consistent with the true values. 
(Detailed comparative results for the power spectra, correlation dimension and three 
types of distance divergences are provided in SI~\cite{SI}.) 

To gain insights into the meaning of the interpretability of the KAN-discovered models, 
we offer a mathematical scheme to interpret machine-learning modeling errors of as 
representing the true underlying system. The issue of considering models that produce 
realistic data, even with orbital errors, is general. In our case, the KAN model 
$\mathbf{G}$ is said to produce identical behavior as the true system $\mathbf{F}$ if 
numerically computed orbits of $\mathbf{G}$ shadow some true orbits of $\mathbf{F}$, at 
least for the observed finite time of the data set. For maps, if a true orbit of 
$\mathbf{F}$ is a sequence 
$\mbox{Orbit}_\mathbf{F}(\mathbf{x}_0)=\{\mathbf{x}_0,\mathbf{F}(\mathbf{x}_0),\mathbf{F}^2(\mathbf{x}_0)...\}\equiv \{\mathbf{x}_0,\mathbf{x}_1,\mathbf{x}_2,...\}$,
it is unreasonable to expect that a good but imperfect model $\mathbf{G}$ will produce an
orbit, denoted as
$\mbox{Orbit}_\mathbf{G}(\mathbf{x}_0)=\{\mathbf{x}_0,\mathbf{G}(\mathbf{x}_0),\mathbf{G}^2(\mathbf{x}_0)...\}\equiv \{\mathbf{x}_0,\tilde{\mathbf{x}}_1,\tilde{\mathbf{x}}_2,...\}$, 
that stays close to $\mbox{Orbit}_F(\mathbf{x}_0)$. If the model is good in 
the sense that a pointwise error $e(x)=|\mathbf{G}(x)-\mathbf{F}(x)|$ on the domain 
$\mathbf{x}\in {\cal D}$ satisfies in terms of the sup-norm, 
$\|e\|_\infty:=\sup_{\mathbf{x}\in {\cal D}} |e(\mathbf{x})|< \epsilon$ for some small 
$\epsilon>0$, then at each step of the model the error is small: 
$\tilde{\mathbf{x}}_{i+1}=\mathbf{G}(\tilde{\mathbf{x}}_i)=F(\tilde{x}_i)+\epsilon_i$ 
and with each step error, $0\leq |\epsilon_i|<\epsilon$. Nonetheless a small normed 
error of the function difference between the system and model alone does not prevent the 
model from producing an unrealistic orbit $\mbox{Orbit}_G(i\mathbf{x}_0)$ that behaves 
quite differently from any orbit of $\mathbf{F}$, e.g., a model orbit that diverges to 
infinity even if the true orbit produces bounded attractor. Furthermore, it is even more 
difficult to consider a model orbit that has statistical properties such as the invariant 
measure of a chaotic attractor analogous to the attractor of the true system.

The KAN was represented as an efficient way to replace a standard multi-layer perceptron 
(MLP)~\cite{LWVRHSHT:2024} and, in so doing, the weights of edges are in principle 
eliminated, but in practice they are absorbed into representing the various activation 
functions at the vertices of the network.  That is, in stating the basic form of a KAN 
as $G(x)=\sum_{q=1}^{2n+1} \Phi_q \circ \sum_{p=1}^n \phi_{q,p}(x_p)$, 
in practice each activation function $\phi_{q,p}$ was represented as a cubic spline 
numerically~\cite{LWVRHSHT:2024}, and therefore each has many internal fitted parameters 
of the scalar piecewise cubics. Collecting all these as the set of parameters 
$\Theta_{q,p}$ for each $\phi_{q,p},$ and $\Theta_q$ for each $\Phi_q$, we can state the 
complete collection of parameters $\Theta=\cup_q (\Theta_q) \cup (\cup_{qp} \Theta_{q,p})$ 
and rewrite the function to emphasize the internal parameters: 
    $G_\Theta(x)=\sum_{q=1}^{2n+1} \Phi_{q,\Theta_q} \circ \sum_{p=1}^n \phi_{q,p,\Theta+{q,p}}(x_p)$,
and for a multivariate argument $x=(x_1,x_2,...,x_d)\in {\mathbb R}^d$. It is 
shown~\cite{LWVRHSHT:2024} that a regularized fit to the data by a loss function 
${\cal L}({\cal D};\Theta)$ (over a data set ${\cal D}$ with respect to the fitting 
parameters $\Theta$), with an objective of data fidelity as least squares fit across the 
data set balanced against $L_2$ norm on the parameters to prevent overfitting.   

While excellent fit when optimizing ${\cal L}({\cal D};\Theta)$ was observed, it is 
possible to emphasize sparsification. That is, one or some of the activation functions 
may be set to zero, a procedure that was called ``pruning''~\cite{LWVRHSHT:2024}. This 
procedure is possible when the representation of the activation functions by splines is
sufficiently fine so that there are more parameters than data points. In such case,  
${\cal L}({\cal D};\Theta)$ will generally have nontrivial level sets. The sparsification 
concept speaks to one of the many reasons to exploit these level sets, generally in terms
of machine-learning interpretability, where the fitted KAN model is pushed toward just a 
few physics recognizable activation functions and the residual in a few terms is collected. 
The mathematical reason this kind of procedure is possible hinges on the implicit function 
theorem~\cite{binmore1982mathematical}. In brief, the KAN model function $G_\Theta(x)$ 
can be varied smoothly with respect to the fitting parameters so that 
${\cal L}({\cal D};\Theta)=c$ is constant for a given parameter $c$.  Therefore even 
following numerical optimization to a small value $c$, there will generally be smooth 
level sets with respect to the $\Theta$ parameters to emphasize other goals of 
explainability. A smooth implicit function $\Theta=h(s)$ exists under the conditions of 
a nonsingular Jacobian derivative $D_\Theta {\cal L}$ that continues a $c$-level set, 
and in principle this level of constancy ${\cal L}({\cal D};h(s))=c$ set may intersect 
the other useful or desirable interpretabile states, including sparsification.

To summarize, we exploited KANs to solve the problem of 
data-driven model discovery for any dynamical systems including those for which the popular 
sparsity-optimization approach to finding the governing equations fails. Our result may 
be understood as realizing shadowing in the functional space where KANs find certain 
functions that produce the same dynamics. These functions may or may not have the same 
mathematical forms as the governing equations of the system and may even be implicit 
with a numerical representation. In the space of all functions, an infinite number of 
such ``shadowing'' functions may exist. We demonstrated that KAN-based machine 
learning can indeed find many of them, depending on the neural-network architecture.

This work was supported by AFOSR under Grant No.~FA9550-21-1-0438. E.B. was
supported by the ONR, ARO, DARPA RSDN, and NIH-NSF CRCNS.


\begin{thebibliography}{59}%
\makeatletter
\providecommand \@ifxundefined [1]{%
 \@ifx{#1\undefined}
}%
\providecommand \@ifnum [1]{%
 \ifnum #1\expandafter \@firstoftwo
 \else \expandafter \@secondoftwo
 \fi
}%
\providecommand \@ifx [1]{%
 \ifx #1\expandafter \@firstoftwo
 \else \expandafter \@secondoftwo
 \fi
}%
\providecommand \natexlab [1]{#1}%
\providecommand \enquote  [1]{``#1''}%
\providecommand \bibnamefont  [1]{#1}%
\providecommand \bibfnamefont [1]{#1}%
\providecommand \citenamefont [1]{#1}%
\providecommand \href@noop [0]{\@secondoftwo}%
\providecommand \href [0]{\begingroup \@sanitize@url \@href}%
\providecommand \@href[1]{\@@startlink{#1}\@@href}%
\providecommand \@@href[1]{\endgroup#1\@@endlink}%
\providecommand \@sanitize@url [0]{\catcode `\\12\catcode `\$12\catcode
  `\&12\catcode `\#12\catcode `\^12\catcode `\_12\catcode `\%12\relax}%
\providecommand \@@startlink[1]{}%
\providecommand \@@endlink[0]{}%
\providecommand \url  [0]{\begingroup\@sanitize@url \@url }%
\providecommand \@url [1]{\endgroup\@href {#1}{\urlprefix }}%
\providecommand \urlprefix  [0]{URL }%
\providecommand \Eprint [0]{\href }%
\providecommand \doibase [0]{https://doi.org/}%
\providecommand \selectlanguage [0]{\@gobble}%
\providecommand \bibinfo  [0]{\@secondoftwo}%
\providecommand \bibfield  [0]{\@secondoftwo}%
\providecommand \translation [1]{[#1]}%
\providecommand \BibitemOpen [0]{}%
\providecommand \bibitemStop [0]{}%
\providecommand \bibitemNoStop [0]{.\EOS\space}%
\providecommand \EOS [0]{\spacefactor3000\relax}%
\providecommand \BibitemShut  [1]{\csname bibitem#1\endcsname}%
\let\auto@bib@innerbib\@empty
\bibitem [{\citenamefont {Farmer}\ and\ \citenamefont
  {Sidorowich}(1987)}]{FS:1987}%
  \BibitemOpen
  \bibfield  {author} {\bibinfo {author} {\bibfnamefont {J.~D.}\ \bibnamefont
  {Farmer}}\ and\ \bibinfo {author} {\bibfnamefont {J.~J.}\ \bibnamefont
  {Sidorowich}},\ }\bibfield  {title} {\bibinfo {title} {Predicting chaotic
  time series},\ }\href {https://doi.org/10.1103/PhysRevLett.59.845} {\bibfield
   {journal} {\bibinfo  {journal} {Phys. Rev. Lett.}\ }\textbf {\bibinfo
  {volume} {59}},\ \bibinfo {pages} {845} (\bibinfo {year} {1987})}\BibitemShut
  {NoStop}%
\bibitem [{\citenamefont {Crutchfield}\ and\ \citenamefont
  {McNamara}(1987)}]{CM:1987}%
  \BibitemOpen
  \bibfield  {author} {\bibinfo {author} {\bibfnamefont {J.~P.}\ \bibnamefont
  {Crutchfield}}\ and\ \bibinfo {author} {\bibfnamefont {B.}~\bibnamefont
  {McNamara}},\ }\bibfield  {title} {\bibinfo {title} {Equations of motion from
  a data series},\ }\href@noop {} {\bibfield  {journal} {\bibinfo  {journal}
  {Complex Sys.}\ }\textbf {\bibinfo {volume} {1}},\ \bibinfo {pages} {417}
  (\bibinfo {year} {1987})}\BibitemShut {NoStop}%
\bibitem [{\citenamefont {Casdagli}(1989)}]{Casdagli:1989}%
  \BibitemOpen
  \bibfield  {author} {\bibinfo {author} {\bibfnamefont {M.}~\bibnamefont
  {Casdagli}},\ }\bibfield  {title} {\bibinfo {title} {Nonlinear prediction of
  chaotic time series},\ }\href@noop {} {\bibfield  {journal} {\bibinfo
  {journal} {Physica D}\ }\textbf {\bibinfo {volume} {35}},\ \bibinfo {pages}
  {335} (\bibinfo {year} {1989})}\BibitemShut {NoStop}%
\bibitem [{\citenamefont {Sugihara}\ \emph {et~al.}(1990)\citenamefont
  {Sugihara}, \citenamefont {Grenfell}, \citenamefont {May}, \citenamefont
  {Chesson}, \citenamefont {Platt},\ and\ \citenamefont
  {Williamson}}]{SGMCPW:1990}%
  \BibitemOpen
  \bibfield  {author} {\bibinfo {author} {\bibfnamefont {G.}~\bibnamefont
  {Sugihara}}, \bibinfo {author} {\bibfnamefont {B.}~\bibnamefont {Grenfell}},
  \bibinfo {author} {\bibfnamefont {R.~M.}\ \bibnamefont {May}}, \bibinfo
  {author} {\bibfnamefont {P.}~\bibnamefont {Chesson}}, \bibinfo {author}
  {\bibfnamefont {H.~M.}\ \bibnamefont {Platt}},\ and\ \bibinfo {author}
  {\bibfnamefont {M.}~\bibnamefont {Williamson}},\ }\bibfield  {title}
  {\bibinfo {title} {Distinguishing error from chaos in ecological time
  series},\ }\href@noop {} {\bibfield  {journal} {\bibinfo  {journal} {Phil.
  Trans. Roy. Soc. London B}\ }\textbf {\bibinfo {volume} {330}},\ \bibinfo
  {pages} {235} (\bibinfo {year} {1990})}\BibitemShut {NoStop}%
\bibitem [{\citenamefont {Kurths}\ and\ \citenamefont
  {Ruzmaikin}(1990)}]{KR:1990}%
  \BibitemOpen
  \bibfield  {author} {\bibinfo {author} {\bibfnamefont {J.}~\bibnamefont
  {Kurths}}\ and\ \bibinfo {author} {\bibfnamefont {A.~A.}\ \bibnamefont
  {Ruzmaikin}},\ }\bibfield  {title} {\bibinfo {title} {On forecasting the
  sunspot numbers},\ }\href@noop {} {\bibfield  {journal} {\bibinfo  {journal}
  {Solar Phys.}\ }\textbf {\bibinfo {volume} {126}},\ \bibinfo {pages} {407}
  (\bibinfo {year} {1990})}\BibitemShut {NoStop}%
\bibitem [{\citenamefont {Grassberger}\ and\ \citenamefont
  {Schreiber}(1990)}]{GS:1990}%
  \BibitemOpen
  \bibfield  {author} {\bibinfo {author} {\bibfnamefont {P.}~\bibnamefont
  {Grassberger}}\ and\ \bibinfo {author} {\bibfnamefont {T.}~\bibnamefont
  {Schreiber}},\ }\bibfield  {title} {\bibinfo {title} {Nonlinear time sequence
  analysis},\ }\href@noop {} {\bibfield  {journal} {\bibinfo  {journal} {Int.
  J. Bif. Chaos}\ }\textbf {\bibinfo {volume} {1}},\ \bibinfo {pages} {521}
  (\bibinfo {year} {1990})}\BibitemShut {NoStop}%
\bibitem [{\citenamefont {Gouesbet}(1991)}]{Gouesbet:1991}%
  \BibitemOpen
  \bibfield  {author} {\bibinfo {author} {\bibfnamefont {G.}~\bibnamefont
  {Gouesbet}},\ }\bibfield  {title} {\bibinfo {title} {{Reconstruction of
  standard and inverse vector fields equivalent to a R\"ossler system}},\
  }\href {https://doi.org/10.1103/PhysRevA.44.6264} {\bibfield  {journal}
  {\bibinfo  {journal} {Phys. Rev. A}\ }\textbf {\bibinfo {volume} {44}},\
  \bibinfo {pages} {6264} (\bibinfo {year} {1991})}\BibitemShut {NoStop}%
\bibitem [{\citenamefont {Tsonis}\ and\ \citenamefont
  {Elsner}(1992)}]{TE:1992}%
  \BibitemOpen
  \bibfield  {author} {\bibinfo {author} {\bibfnamefont {A.~A.}\ \bibnamefont
  {Tsonis}}\ and\ \bibinfo {author} {\bibfnamefont {J.~B.}\ \bibnamefont
  {Elsner}},\ }\bibfield  {title} {\bibinfo {title} {Nonlinear prediction as a
  way of distinguishing chaos from random fractal sequences},\ }\href@noop {}
  {\bibfield  {journal} {\bibinfo  {journal} {Nature (London)}\ }\textbf
  {\bibinfo {volume} {358}},\ \bibinfo {pages} {217} (\bibinfo {year}
  {1992})}\BibitemShut {NoStop}%
\bibitem [{\citenamefont {Baake}\ \emph {et~al.}(1992)\citenamefont {Baake},
  \citenamefont {Baake}, \citenamefont {Bock},\ and\ \citenamefont
  {Briggs}}]{BBBB:1992}%
  \BibitemOpen
  \bibfield  {author} {\bibinfo {author} {\bibfnamefont {E.}~\bibnamefont
  {Baake}}, \bibinfo {author} {\bibfnamefont {M.}~\bibnamefont {Baake}},
  \bibinfo {author} {\bibfnamefont {H.~G.}\ \bibnamefont {Bock}},\ and\
  \bibinfo {author} {\bibfnamefont {K.~M.}\ \bibnamefont {Briggs}},\ }\bibfield
   {title} {\bibinfo {title} {Fitting ordinary differential equations to
  chaotic data},\ }\href {https://doi.org/10.1103/PhysRevA.45.5524} {\bibfield
  {journal} {\bibinfo  {journal} {Phys. Rev. A}\ }\textbf {\bibinfo {volume}
  {45}},\ \bibinfo {pages} {5524} (\bibinfo {year} {1992})}\BibitemShut
  {NoStop}%
\bibitem [{\citenamefont {Longtin}(1993)}]{Longtin:1993}%
  \BibitemOpen
  \bibfield  {author} {\bibinfo {author} {\bibfnamefont {A.}~\bibnamefont
  {Longtin}},\ }\bibfield  {title} {\bibinfo {title} {Nonlinear forecasting of
  spike trains from sensory neurons},\ }\href@noop {} {\bibfield  {journal}
  {\bibinfo  {journal} {Int. J. Bif. Chaos}\ }\textbf {\bibinfo {volume} {3}},\
  \bibinfo {pages} {651} (\bibinfo {year} {1993})}\BibitemShut {NoStop}%
\bibitem [{\citenamefont {Murray}(1993)}]{Murray:1993}%
  \BibitemOpen
  \bibfield  {author} {\bibinfo {author} {\bibfnamefont {D.~B.}\ \bibnamefont
  {Murray}},\ }\bibfield  {title} {\bibinfo {title} {Forecasting a chaotic time
  series using an improved metric for embedding space},\ }\href@noop {}
  {\bibfield  {journal} {\bibinfo  {journal} {Physica D}\ }\textbf {\bibinfo
  {volume} {68}},\ \bibinfo {pages} {318} (\bibinfo {year} {1993})}\BibitemShut
  {NoStop}%
\bibitem [{\citenamefont {Sauer}(1994)}]{Sauer:1994}%
  \BibitemOpen
  \bibfield  {author} {\bibinfo {author} {\bibfnamefont {T.}~\bibnamefont
  {Sauer}},\ }\bibfield  {title} {\bibinfo {title} {Reconstruction of dynamical
  systems from interspike intervals},\ }\href
  {https://doi.org/10.1103/PhysRevLett.72.3811} {\bibfield  {journal} {\bibinfo
   {journal} {Phys. Rev. Lett.}\ }\textbf {\bibinfo {volume} {72}},\ \bibinfo
  {pages} {3811} (\bibinfo {year} {1994})}\BibitemShut {NoStop}%
\bibitem [{\citenamefont {Sugihara}(1994)}]{Sugihara:1994}%
  \BibitemOpen
  \bibfield  {author} {\bibinfo {author} {\bibfnamefont {G.}~\bibnamefont
  {Sugihara}},\ }\bibfield  {title} {\bibinfo {title} {Nonlinear forecasting
  for the classification of natural time series},\ }\href@noop {} {\bibfield
  {journal} {\bibinfo  {journal} {Philos. T. Roy. Soc. A.}\ }\textbf {\bibinfo
  {volume} {348}},\ \bibinfo {pages} {477} (\bibinfo {year}
  {1994})}\BibitemShut {NoStop}%
\bibitem [{\citenamefont {Finkenst\"{a}dt}\ and\ \citenamefont
  {Kuhbier}(1995)}]{FK:1995}%
  \BibitemOpen
  \bibfield  {author} {\bibinfo {author} {\bibfnamefont {B.}~\bibnamefont
  {Finkenst\"{a}dt}}\ and\ \bibinfo {author} {\bibfnamefont {P.}~\bibnamefont
  {Kuhbier}},\ }\bibfield  {title} {\bibinfo {title} {Forecasting nonlinear
  economic time series: A simple test to accompany the nearest neighbor
  approach},\ }\href@noop {} {\bibfield  {journal} {\bibinfo  {journal}
  {Empiri. Econ.}\ }\textbf {\bibinfo {volume} {20}},\ \bibinfo {pages} {243}
  (\bibinfo {year} {1995})}\BibitemShut {NoStop}%
\bibitem [{\citenamefont {Parlitz}(1996)}]{Parlitz:1996}%
  \BibitemOpen
  \bibfield  {author} {\bibinfo {author} {\bibfnamefont {U.}~\bibnamefont
  {Parlitz}},\ }\bibfield  {title} {\bibinfo {title} {Estimating model
  parameters from time series by autosynchronization},\ }\href@noop {}
  {\bibfield  {journal} {\bibinfo  {journal} {Phys. Rev. Lett.}\ }\textbf
  {\bibinfo {volume} {76}},\ \bibinfo {pages} {1232} (\bibinfo {year}
  {1996})}\BibitemShut {NoStop}%
\bibitem [{\citenamefont {Schiff}\ \emph {et~al.}(1996)\citenamefont {Schiff},
  \citenamefont {So}, \citenamefont {Chang}, \citenamefont {Burke},\ and\
  \citenamefont {Sauer}}]{SSCBS:1996}%
  \BibitemOpen
  \bibfield  {author} {\bibinfo {author} {\bibfnamefont {S.~J.}\ \bibnamefont
  {Schiff}}, \bibinfo {author} {\bibfnamefont {P.}~\bibnamefont {So}}, \bibinfo
  {author} {\bibfnamefont {T.}~\bibnamefont {Chang}}, \bibinfo {author}
  {\bibfnamefont {R.~E.}\ \bibnamefont {Burke}},\ and\ \bibinfo {author}
  {\bibfnamefont {T.}~\bibnamefont {Sauer}},\ }\bibfield  {title} {\bibinfo
  {title} {Detecting dynamical interdependence and generalized synchrony
  through mutual prediction in a neural ensemble},\ }\href
  {https://doi.org/10.1103/PhysRevE.54.6708} {\bibfield  {journal} {\bibinfo
  {journal} {Phys. Rev. E}\ }\textbf {\bibinfo {volume} {54}},\ \bibinfo
  {pages} {6708} (\bibinfo {year} {1996})}\BibitemShut {NoStop}%
\bibitem [{\citenamefont {Szpiro}(1997)}]{Szpiro:1997}%
  \BibitemOpen
  \bibfield  {author} {\bibinfo {author} {\bibfnamefont {G.~G.}\ \bibnamefont
  {Szpiro}},\ }\bibfield  {title} {\bibinfo {title} {Forecasting chaotic time
  series with genetic algorithms},\ }\href
  {https://doi.org/10.1103/PhysRevE.55.2557} {\bibfield  {journal} {\bibinfo
  {journal} {Phys. Rev. E}\ }\textbf {\bibinfo {volume} {55}},\ \bibinfo
  {pages} {2557} (\bibinfo {year} {1997})}\BibitemShut {NoStop}%
\bibitem [{\citenamefont {Hegger}\ \emph {et~al.}(1999)\citenamefont {Hegger},
  \citenamefont {Kantz},\ and\ \citenamefont {Schreiber}}]{HKS:1999}%
  \BibitemOpen
  \bibfield  {author} {\bibinfo {author} {\bibfnamefont {R.}~\bibnamefont
  {Hegger}}, \bibinfo {author} {\bibfnamefont {H.}~\bibnamefont {Kantz}},\ and\
  \bibinfo {author} {\bibfnamefont {T.}~\bibnamefont {Schreiber}},\ }\bibfield
  {title} {\bibinfo {title} {Practical implementation of nonlinear time series
  methods: The tisean package},\ }\href@noop {} {\bibfield  {journal} {\bibinfo
   {journal} {Chaos}\ }\textbf {\bibinfo {volume} {9}},\ \bibinfo {pages} {413}
  (\bibinfo {year} {1999})}\BibitemShut {NoStop}%
\bibitem [{\citenamefont {Bollt}(2000)}]{Bollt:2000}%
  \BibitemOpen
  \bibfield  {author} {\bibinfo {author} {\bibfnamefont {E.~M.}\ \bibnamefont
  {Bollt}},\ }\bibfield  {title} {\bibinfo {title} {Controlling chaos and the
  inverse frobenius-perron problem: global stabilization of arbitrary invariant
  measures},\ }\href@noop {} {\bibfield  {journal} {\bibinfo  {journal} {Int.
  J. Bif. Chaos}\ }\textbf {\bibinfo {volume} {10}},\ \bibinfo {pages} {1033}
  (\bibinfo {year} {2000})}\BibitemShut {NoStop}%
\bibitem [{\citenamefont {Hegger}\ \emph {et~al.}(2000)\citenamefont {Hegger},
  \citenamefont {Kantz}, \citenamefont {Matassini},\ and\ \citenamefont
  {Schreiber}}]{HKMS:2000}%
  \BibitemOpen
  \bibfield  {author} {\bibinfo {author} {\bibfnamefont {R.}~\bibnamefont
  {Hegger}}, \bibinfo {author} {\bibfnamefont {H.}~\bibnamefont {Kantz}},
  \bibinfo {author} {\bibfnamefont {L.}~\bibnamefont {Matassini}},\ and\
  \bibinfo {author} {\bibfnamefont {T.}~\bibnamefont {Schreiber}},\ }\bibfield
  {title} {\bibinfo {title} {Coping with nonstationarity by overembedding},\
  }\href {https://doi.org/10.1103/PhysRevLett.84.4092} {\bibfield  {journal}
  {\bibinfo  {journal} {Phys. Rev. Lett.}\ }\textbf {\bibinfo {volume} {84}},\
  \bibinfo {pages} {4092} (\bibinfo {year} {2000})}\BibitemShut {NoStop}%
\bibitem [{\citenamefont {Sello}(2001)}]{Sello:2001}%
  \BibitemOpen
  \bibfield  {author} {\bibinfo {author} {\bibfnamefont {S.}~\bibnamefont
  {Sello}},\ }\bibfield  {title} {\bibinfo {title} {Solar cycle forecasting: a
  nonlinear dynamics approach},\ }\href@noop {} {\bibfield  {journal} {\bibinfo
   {journal} {Astron. Astrophys.}\ }\textbf {\bibinfo {volume} {377}},\
  \bibinfo {pages} {312} (\bibinfo {year} {2001})}\BibitemShut {NoStop}%
\bibitem [{\citenamefont {Matsumoto}\ \emph {et~al.}(2001)\citenamefont
  {Matsumoto}, \citenamefont {Nakajima}, \citenamefont {Saito}, \citenamefont
  {Sugi},\ and\ \citenamefont {Hamagishi}}]{MNSSH:2001}%
  \BibitemOpen
  \bibfield  {author} {\bibinfo {author} {\bibfnamefont {T.}~\bibnamefont
  {Matsumoto}}, \bibinfo {author} {\bibfnamefont {Y.}~\bibnamefont {Nakajima}},
  \bibinfo {author} {\bibfnamefont {M.}~\bibnamefont {Saito}}, \bibinfo
  {author} {\bibfnamefont {J.}~\bibnamefont {Sugi}},\ and\ \bibinfo {author}
  {\bibfnamefont {H.}~\bibnamefont {Hamagishi}},\ }\bibfield  {title} {\bibinfo
  {title} {Reconstructions and predictions of nonlinear dynamical systems: a
  hierarchical bayesian approach},\ }\href@noop {} {\bibfield  {journal}
  {\bibinfo  {journal} {IEEE Trans. Signal Proc.}\ }\textbf {\bibinfo {volume}
  {49}},\ \bibinfo {pages} {2138} (\bibinfo {year} {2001})}\BibitemShut
  {NoStop}%
\bibitem [{\citenamefont {Smith}(2002)}]{Smith:2002}%
  \BibitemOpen
  \bibfield  {author} {\bibinfo {author} {\bibfnamefont {L.~A.}\ \bibnamefont
  {Smith}},\ }\bibfield  {title} {\bibinfo {title} {What might we learn from
  climate forecasts?},\ }\href@noop {} {\bibfield  {journal} {\bibinfo
  {journal} {Proc. Nat. Acad. Sci. (USA)}\ }\textbf {\bibinfo {volume} {19}},\
  \bibinfo {pages} {2487} (\bibinfo {year} {2002})}\BibitemShut {NoStop}%
\bibitem [{\citenamefont {Judd}(2003)}]{Judd:2003}%
  \BibitemOpen
  \bibfield  {author} {\bibinfo {author} {\bibfnamefont {K.}~\bibnamefont
  {Judd}},\ }\bibfield  {title} {\bibinfo {title} {Nonlinear state estimation,
  indistinguishable states, and the extended kalman filter},\ }\href@noop {}
  {\bibfield  {journal} {\bibinfo  {journal} {Physica D}\ }\textbf {\bibinfo
  {volume} {183}},\ \bibinfo {pages} {273} (\bibinfo {year}
  {2003})}\BibitemShut {NoStop}%
\bibitem [{\citenamefont {Sauer}(2004)}]{Sauer:2004}%
  \BibitemOpen
  \bibfield  {author} {\bibinfo {author} {\bibfnamefont {T.~D.}\ \bibnamefont
  {Sauer}},\ }\bibfield  {title} {\bibinfo {title} {Reconstruction of shared
  nonlinear dynamics in a network},\ }\href
  {https://doi.org/10.1103/PhysRevLett.93.198701} {\bibfield  {journal}
  {\bibinfo  {journal} {Phys. Rev. Lett.}\ }\textbf {\bibinfo {volume} {93}},\
  \bibinfo {pages} {198701} (\bibinfo {year} {2004})}\BibitemShut {NoStop}%
\bibitem [{\citenamefont {Yao}\ and\ \citenamefont {Bollt}(2007)}]{YB:2007}%
  \BibitemOpen
  \bibfield  {author} {\bibinfo {author} {\bibfnamefont {C.}~\bibnamefont
  {Yao}}\ and\ \bibinfo {author} {\bibfnamefont {E.~M.}\ \bibnamefont
  {Bollt}},\ }\bibfield  {title} {\bibinfo {title} {Modeling and nonlinear
  parameter estimation with {Kronecker} product representation for coupled
  oscillators and spatiotemporal systems},\ }\href@noop {} {\bibfield
  {journal} {\bibinfo  {journal} {Physica D}\ }\textbf {\bibinfo {volume}
  {227}},\ \bibinfo {pages} {78} (\bibinfo {year} {2007})}\BibitemShut
  {NoStop}%
\bibitem [{\citenamefont {Tao}\ \emph {et~al.}(2007)\citenamefont {Tao},
  \citenamefont {Zhang},\ and\ \citenamefont {Jiang}}]{TZJ:2007}%
  \BibitemOpen
  \bibfield  {author} {\bibinfo {author} {\bibfnamefont {C.}~\bibnamefont
  {Tao}}, \bibinfo {author} {\bibfnamefont {Y.}~\bibnamefont {Zhang}},\ and\
  \bibinfo {author} {\bibfnamefont {J.~J.}\ \bibnamefont {Jiang}},\ }\bibfield
  {title} {\bibinfo {title} {Estimating system parameters from chaotic time
  series with synchronization optimized by a genetic algorithm},\ }\href
  {https://doi.org/10.1103/PhysRevE.76.016209} {\bibfield  {journal} {\bibinfo
  {journal} {Phys. Rev. E}\ }\textbf {\bibinfo {volume} {76}},\ \bibinfo
  {pages} {016209} (\bibinfo {year} {2007})}\BibitemShut {NoStop}%
\bibitem [{\citenamefont {Wang}\ \emph
  {et~al.}(2011{\natexlab{a}})\citenamefont {Wang}, \citenamefont {Yang},
  \citenamefont {Lai}, \citenamefont {Kovanis},\ and\ \citenamefont
  {Grebogi}}]{WYLKG:2011}%
  \BibitemOpen
  \bibfield  {author} {\bibinfo {author} {\bibfnamefont {W.-X.}\ \bibnamefont
  {Wang}}, \bibinfo {author} {\bibfnamefont {R.}~\bibnamefont {Yang}}, \bibinfo
  {author} {\bibfnamefont {Y.-C.}\ \bibnamefont {Lai}}, \bibinfo {author}
  {\bibfnamefont {V.}~\bibnamefont {Kovanis}},\ and\ \bibinfo {author}
  {\bibfnamefont {C.}~\bibnamefont {Grebogi}},\ }\bibfield  {title} {\bibinfo
  {title} {Predicting catastrophes in nonlinear dynamical systems by
  compressive sensing},\ }\href@noop {} {\bibfield  {journal} {\bibinfo
  {journal} {Phys. Rev. Lett.}\ }\textbf {\bibinfo {volume} {106}},\ \bibinfo
  {pages} {154101} (\bibinfo {year} {2011}{\natexlab{a}})}\BibitemShut
  {NoStop}%
\bibitem [{\citenamefont {Wang}\ \emph
  {et~al.}(2011{\natexlab{b}})\citenamefont {Wang}, \citenamefont {Lai},
  \citenamefont {Grebogi},\ and\ \citenamefont {Ye}}]{WLGY:2011}%
  \BibitemOpen
  \bibfield  {author} {\bibinfo {author} {\bibfnamefont {W.-X.}\ \bibnamefont
  {Wang}}, \bibinfo {author} {\bibfnamefont {Y.-C.}\ \bibnamefont {Lai}},
  \bibinfo {author} {\bibfnamefont {C.}~\bibnamefont {Grebogi}},\ and\ \bibinfo
  {author} {\bibfnamefont {J.-P.}\ \bibnamefont {Ye}},\ }\bibfield  {title}
  {\bibinfo {title} {Network reconstruction based on evolutionary-game data via
  compressive sensing},\ }\href@noop {} {\bibfield  {journal} {\bibinfo
  {journal} {Phys. Rev. X}\ }\textbf {\bibinfo {volume} {1}},\ \bibinfo {pages}
  {021021} (\bibinfo {year} {2011}{\natexlab{b}})}\BibitemShut {NoStop}%
\bibitem [{\citenamefont {Wang}\ \emph
  {et~al.}(2011{\natexlab{c}})\citenamefont {Wang}, \citenamefont {Yang},
  \citenamefont {Lai}, \citenamefont {Kovanis},\ and\ \citenamefont
  {Harrison}}]{WYLKH:2011}%
  \BibitemOpen
  \bibfield  {author} {\bibinfo {author} {\bibfnamefont {W.-X.}\ \bibnamefont
  {Wang}}, \bibinfo {author} {\bibfnamefont {R.}~\bibnamefont {Yang}}, \bibinfo
  {author} {\bibfnamefont {Y.-C.}\ \bibnamefont {Lai}}, \bibinfo {author}
  {\bibfnamefont {V.}~\bibnamefont {Kovanis}},\ and\ \bibinfo {author}
  {\bibfnamefont {M.~A.~F.}\ \bibnamefont {Harrison}},\ }\bibfield  {title}
  {\bibinfo {title} {Time-series-based prediction of complex oscillator
  networks via compressive sensing},\ }\href@noop {} {\bibfield  {journal}
  {\bibinfo  {journal} {EPL (Europhys. Lett.)}\ }\textbf {\bibinfo {volume}
  {94}},\ \bibinfo {pages} {48006} (\bibinfo {year}
  {2011}{\natexlab{c}})}\BibitemShut {NoStop}%
\bibitem [{\citenamefont {Su}\ \emph {et~al.}(2012{\natexlab{a}})\citenamefont
  {Su}, \citenamefont {Ni}, \citenamefont {Wang},\ and\ \citenamefont
  {Lai}}]{SNWL:2012}%
  \BibitemOpen
  \bibfield  {author} {\bibinfo {author} {\bibfnamefont {R.-Q.}\ \bibnamefont
  {Su}}, \bibinfo {author} {\bibfnamefont {X.}~\bibnamefont {Ni}}, \bibinfo
  {author} {\bibfnamefont {W.-X.}\ \bibnamefont {Wang}},\ and\ \bibinfo
  {author} {\bibfnamefont {Y.-C.}\ \bibnamefont {Lai}},\ }\bibfield  {title}
  {\bibinfo {title} {Forecasting synchronizability of complex networks from
  data},\ }\href@noop {} {\bibfield  {journal} {\bibinfo  {journal} {Phys. Rev.
  E}\ }\textbf {\bibinfo {volume} {85}},\ \bibinfo {pages} {056220} (\bibinfo
  {year} {2012}{\natexlab{a}})}\BibitemShut {NoStop}%
\bibitem [{\citenamefont {Su}\ \emph {et~al.}(2012{\natexlab{b}})\citenamefont
  {Su}, \citenamefont {Wang},\ and\ \citenamefont {Lai}}]{SWL:2012}%
  \BibitemOpen
  \bibfield  {author} {\bibinfo {author} {\bibfnamefont {R.-Q.}\ \bibnamefont
  {Su}}, \bibinfo {author} {\bibfnamefont {W.-X.}\ \bibnamefont {Wang}},\ and\
  \bibinfo {author} {\bibfnamefont {Y.-C.}\ \bibnamefont {Lai}},\ }\bibfield
  {title} {\bibinfo {title} {Detecting hidden nodes in complex networks from
  time series},\ }\href@noop {} {\bibfield  {journal} {\bibinfo  {journal}
  {Phys. Rev. E}\ }\textbf {\bibinfo {volume} {85}},\ \bibinfo {pages} {065201}
  (\bibinfo {year} {2012}{\natexlab{b}})}\BibitemShut {NoStop}%
\bibitem [{\citenamefont {Su}\ \emph {et~al.}(2014{\natexlab{a}})\citenamefont
  {Su}, \citenamefont {Lai},\ and\ \citenamefont {Wang}}]{SWL:2014}%
  \BibitemOpen
  \bibfield  {author} {\bibinfo {author} {\bibfnamefont {R.-Q.}\ \bibnamefont
  {Su}}, \bibinfo {author} {\bibfnamefont {Y.-C.}\ \bibnamefont {Lai}},\ and\
  \bibinfo {author} {\bibfnamefont {X.}~\bibnamefont {Wang}},\ }\bibfield
  {title} {\bibinfo {title} {Identifying chaotic fitzhugh-nagumo neurons using
  compressive sensing},\ }\href@noop {} {\bibfield  {journal} {\bibinfo
  {journal} {Entropy}\ }\textbf {\bibinfo {volume} {16}},\ \bibinfo {pages}
  {3889} (\bibinfo {year} {2014}{\natexlab{a}})}\BibitemShut {NoStop}%
\bibitem [{\citenamefont {Su}\ \emph {et~al.}(2014{\natexlab{b}})\citenamefont
  {Su}, \citenamefont {Lai}, \citenamefont {Wang},\ and\ \citenamefont
  {Do}}]{SLWD:2014}%
  \BibitemOpen
  \bibfield  {author} {\bibinfo {author} {\bibfnamefont {R.-Q.}\ \bibnamefont
  {Su}}, \bibinfo {author} {\bibfnamefont {Y.-C.}\ \bibnamefont {Lai}},
  \bibinfo {author} {\bibfnamefont {X.}~\bibnamefont {Wang}},\ and\ \bibinfo
  {author} {\bibfnamefont {Y.-H.}\ \bibnamefont {Do}},\ }\bibfield  {title}
  {\bibinfo {title} {Uncovering hidden nodes in complex networks in the
  presence of noise},\ }\href@noop {} {\bibfield  {journal} {\bibinfo
  {journal} {Sci. Rep.}\ }\textbf {\bibinfo {volume} {4}},\ \bibinfo {pages}
  {3944} (\bibinfo {year} {2014}{\natexlab{b}})}\BibitemShut {NoStop}%
\bibitem [{\citenamefont {Shen}\ \emph {et~al.}(2014)\citenamefont {Shen},
  \citenamefont {Wang}, \citenamefont {Fan}, \citenamefont {Di},\ and\
  \citenamefont {Lai}}]{SWFDL:2014}%
  \BibitemOpen
  \bibfield  {author} {\bibinfo {author} {\bibfnamefont {Z.}~\bibnamefont
  {Shen}}, \bibinfo {author} {\bibfnamefont {W.-X.}\ \bibnamefont {Wang}},
  \bibinfo {author} {\bibfnamefont {Y.}~\bibnamefont {Fan}}, \bibinfo {author}
  {\bibfnamefont {Z.}~\bibnamefont {Di}},\ and\ \bibinfo {author}
  {\bibfnamefont {Y.-C.}\ \bibnamefont {Lai}},\ }\bibfield  {title} {\bibinfo
  {title} {Reconstructing propagation networks with natural diversity and
  identifying hidden sources},\ }\href@noop {} {\bibfield  {journal} {\bibinfo
  {journal} {Nat. Commun.}\ }\textbf {\bibinfo {volume} {5}},\ \bibinfo {pages}
  {4323} (\bibinfo {year} {2014})}\BibitemShut {NoStop}%
\bibitem [{\citenamefont {Su}\ \emph {et~al.}(2016)\citenamefont {Su},
  \citenamefont {Wang}, \citenamefont {Wang},\ and\ \citenamefont
  {Lai}}]{SWWL:2016}%
  \BibitemOpen
  \bibfield  {author} {\bibinfo {author} {\bibfnamefont {R.-Q.}\ \bibnamefont
  {Su}}, \bibinfo {author} {\bibfnamefont {W.-W.}\ \bibnamefont {Wang}},
  \bibinfo {author} {\bibfnamefont {X.}~\bibnamefont {Wang}},\ and\ \bibinfo
  {author} {\bibfnamefont {Y.-C.}\ \bibnamefont {Lai}},\ }\bibfield  {title}
  {\bibinfo {title} {Data based reconstruction of complex geospatial networks,
  nodal positioning, and detection of hidden node},\ }\href@noop {} {\bibfield
  {journal} {\bibinfo  {journal} {R. Soc. Open Sci.}\ }\textbf {\bibinfo
  {volume} {3}},\ \bibinfo {pages} {150577} (\bibinfo {year}
  {2016})}\BibitemShut {NoStop}%
\bibitem [{\citenamefont {AlMomani}\ \emph {et~al.}(2020)\citenamefont
  {AlMomani}, \citenamefont {Jie},\ and\ \citenamefont {Bollt}}]{ASB:2020}%
  \BibitemOpen
  \bibfield  {author} {\bibinfo {author} {\bibfnamefont {A.~A.~R.}\
  \bibnamefont {AlMomani}}, \bibinfo {author} {\bibfnamefont {S.}~\bibnamefont
  {Jie}},\ and\ \bibinfo {author} {\bibfnamefont {E.~M.}\ \bibnamefont
  {Bollt}},\ }\bibfield  {title} {\bibinfo {title} {How entropic regression
  beats the outliers problem in nonlinear system identification},\ }\href@noop
  {} {\bibfield  {journal} {\bibinfo  {journal} {Chaos}\ }\textbf {\bibinfo
  {volume} {30}},\ \bibinfo {pages} {013107} (\bibinfo {year}
  {2020})}\BibitemShut {NoStop}%
\bibitem [{\citenamefont {Yang}\ \emph {et~al.}(2012)\citenamefont {Yang},
  \citenamefont {Lai},\ and\ \citenamefont {Grebogi}}]{YLG:2012}%
  \BibitemOpen
  \bibfield  {author} {\bibinfo {author} {\bibfnamefont {R.}~\bibnamefont
  {Yang}}, \bibinfo {author} {\bibfnamefont {Y.-C.}\ \bibnamefont {Lai}},\ and\
  \bibinfo {author} {\bibfnamefont {C.}~\bibnamefont {Grebogi}},\ }\bibfield
  {title} {\bibinfo {title} {Forecasting the future: is it possible for
  time-varying nonlinear dynamical systems?},\ }\href@noop {} {\bibfield
  {journal} {\bibinfo  {journal} {Chaos}\ }\textbf {\bibinfo {volume} {22}},\
  \bibinfo {pages} {033119} (\bibinfo {year} {2012})}\BibitemShut {NoStop}%
\bibitem [{\citenamefont {Cand\`{e}s}\ \emph
  {et~al.}(2006{\natexlab{a}})\citenamefont {Cand\`{e}s}, \citenamefont
  {Romberg},\ and\ \citenamefont {Tao}}]{CRT:2006a}%
  \BibitemOpen
  \bibfield  {author} {\bibinfo {author} {\bibfnamefont {E.}~\bibnamefont
  {Cand\`{e}s}}, \bibinfo {author} {\bibfnamefont {J.}~\bibnamefont
  {Romberg}},\ and\ \bibinfo {author} {\bibfnamefont {T.}~\bibnamefont {Tao}},\
  }\bibfield  {title} {\bibinfo {title} {Robust uncertainty principles: exact
  signal reconstruction from highly incomplete frequency information},\
  }\href@noop {} {\bibfield  {journal} {\bibinfo  {journal} {IEEE Trans. Info.
  Theory}\ }\textbf {\bibinfo {volume} {52}},\ \bibinfo {pages} {489} (\bibinfo
  {year} {2006}{\natexlab{a}})}\BibitemShut {NoStop}%
\bibitem [{\citenamefont {Cand\`{e}s}\ \emph
  {et~al.}(2006{\natexlab{b}})\citenamefont {Cand\`{e}s}, \citenamefont
  {Romberg},\ and\ \citenamefont {Tao}}]{CRT:2006b}%
  \BibitemOpen
  \bibfield  {author} {\bibinfo {author} {\bibfnamefont {E.}~\bibnamefont
  {Cand\`{e}s}}, \bibinfo {author} {\bibfnamefont {J.}~\bibnamefont
  {Romberg}},\ and\ \bibinfo {author} {\bibfnamefont {T.}~\bibnamefont {Tao}},\
  }\bibfield  {title} {\bibinfo {title} {Stable signal recovery from incomplete
  and inaccurate measurements},\ }\href@noop {} {\bibfield  {journal} {\bibinfo
   {journal} {Comm. Pure Appl. Math.}\ }\textbf {\bibinfo {volume} {59}},\
  \bibinfo {pages} {1207} (\bibinfo {year} {2006}{\natexlab{b}})}\BibitemShut
  {NoStop}%
\bibitem [{\citenamefont {Donoho}(2006)}]{Donoho:2006}%
  \BibitemOpen
  \bibfield  {author} {\bibinfo {author} {\bibfnamefont {D.}~\bibnamefont
  {Donoho}},\ }\bibfield  {title} {\bibinfo {title} {Compressed sensing},\
  }\href@noop {} {\bibfield  {journal} {\bibinfo  {journal} {IEEE Trans. Info.
  Theory}\ }\textbf {\bibinfo {volume} {52}},\ \bibinfo {pages} {1289}
  (\bibinfo {year} {2006})}\BibitemShut {NoStop}%
\bibitem [{\citenamefont {Baraniuk}(2007)}]{Baraniuk:2007}%
  \BibitemOpen
  \bibfield  {author} {\bibinfo {author} {\bibfnamefont {R.~G.}\ \bibnamefont
  {Baraniuk}},\ }\bibfield  {title} {\bibinfo {title} {Compressed sensing},\
  }\href@noop {} {\bibfield  {journal} {\bibinfo  {journal} {IEEE Signal
  Process. Mag.}\ }\textbf {\bibinfo {volume} {24}},\ \bibinfo {pages} {118}
  (\bibinfo {year} {2007})}\BibitemShut {NoStop}%
\bibitem [{\citenamefont {Cande\`s}\ and\ \citenamefont
  {Wakin}(2008)}]{CW:2008}%
  \BibitemOpen
  \bibfield  {author} {\bibinfo {author} {\bibfnamefont {E.}~\bibnamefont
  {Cande\`s}}\ and\ \bibinfo {author} {\bibfnamefont {M.}~\bibnamefont
  {Wakin}},\ }\bibfield  {title} {\bibinfo {title} {An introduction to
  compressive sampling},\ }\href@noop {} {\bibfield  {journal} {\bibinfo
  {journal} {IEEE Signal Process. Mag.}\ }\textbf {\bibinfo {volume} {25}},\
  \bibinfo {pages} {21} (\bibinfo {year} {2008})}\BibitemShut {NoStop}%
\bibitem [{\citenamefont {Brunton}\ \emph {et~al.}(2016)\citenamefont
  {Brunton}, \citenamefont {Proctor},\ and\ \citenamefont {Kutz}}]{BPK:2016}%
  \BibitemOpen
  \bibfield  {author} {\bibinfo {author} {\bibfnamefont {S.~L.}\ \bibnamefont
  {Brunton}}, \bibinfo {author} {\bibfnamefont {J.~L.}\ \bibnamefont
  {Proctor}},\ and\ \bibinfo {author} {\bibfnamefont {J.~N.}\ \bibnamefont
  {Kutz}},\ }\bibfield  {title} {\bibinfo {title} {Discovering governing
  equations from data by sparse identification of nonlinear dynamical
  systems},\ }\href@noop {} {\bibfield  {journal} {\bibinfo  {journal} {Proc.
  Nat. Acad. Sci. (USA)}\ }\textbf {\bibinfo {volume} {113}},\ \bibinfo {pages}
  {3932} (\bibinfo {year} {2016})}\BibitemShut {NoStop}%
\bibitem [{\citenamefont {Lai}(2021)}]{Lai:2021}%
  \BibitemOpen
  \bibfield  {author} {\bibinfo {author} {\bibfnamefont {Y.-C.}\ \bibnamefont
  {Lai}},\ }\bibfield  {title} {\bibinfo {title} {Finding nonlinear system
  equations and complex network structures from data: {A} sparse optimization
  approach},\ }\href@noop {} {\bibfield  {journal} {\bibinfo  {journal}
  {Chaos}\ }\textbf {\bibinfo {volume} {31}},\ \bibinfo {pages} {082101}
  (\bibinfo {year} {2021})}\BibitemShut {NoStop}%
\bibitem [{\citenamefont {Lorenz}(1963)}]{Lorenz:1963}%
  \BibitemOpen
  \bibfield  {author} {\bibinfo {author} {\bibfnamefont {E.~N.}\ \bibnamefont
  {Lorenz}},\ }\bibfield  {title} {\bibinfo {title} {Deterministic nonperiodic
  flow},\ }\href@noop {} {\bibfield  {journal} {\bibinfo  {journal} {J. Atmos.
  Sci.}\ }\textbf {\bibinfo {volume} {20}},\ \bibinfo {pages} {130} (\bibinfo
  {year} {1963})}\BibitemShut {NoStop}%
\bibitem [{\citenamefont {R\"{o}ssler}(1976)}]{Rossler:1976}%
  \BibitemOpen
  \bibfield  {author} {\bibinfo {author} {\bibfnamefont {O.~E.}\ \bibnamefont
  {R\"{o}ssler}},\ }\bibfield  {title} {\bibinfo {title} {Equation for
  continuous chaos},\ }\href@noop {} {\bibfield  {journal} {\bibinfo  {journal}
  {Phys. Lett. A}\ }\textbf {\bibinfo {volume} {57}},\ \bibinfo {pages} {397}
  (\bibinfo {year} {1976})}\BibitemShut {NoStop}%
\bibitem [{\citenamefont {Ikeda}(1979)}]{Ikeda:1979}%
  \BibitemOpen
  \bibfield  {author} {\bibinfo {author} {\bibfnamefont {K.}~\bibnamefont
  {Ikeda}},\ }\bibfield  {title} {\bibinfo {title} {Multiple-valued stationary
  state and its instability of the transmitted light by a ring cavity system},\
  }\href@noop {} {\bibfield  {journal} {\bibinfo  {journal} {Opt. Commun.}\
  }\textbf {\bibinfo {volume} {30}},\ \bibinfo {pages} {257} (\bibinfo {year}
  {1979})}\BibitemShut {NoStop}%
\bibitem [{\citenamefont {Hammel}\ \emph {et~al.}(1985)\citenamefont {Hammel},
  \citenamefont {Jones},\ and\ \citenamefont {Moloney}}]{HJM:1985}%
  \BibitemOpen
  \bibfield  {author} {\bibinfo {author} {\bibfnamefont {S.~M.}\ \bibnamefont
  {Hammel}}, \bibinfo {author} {\bibfnamefont {C.~K. R.~T.}\ \bibnamefont
  {Jones}},\ and\ \bibinfo {author} {\bibfnamefont {J.~V.}\ \bibnamefont
  {Moloney}},\ }\bibfield  {title} {\bibinfo {title} {Global dynamical behavior
  of the optical field in a ring cavity},\ }\href@noop {} {\bibfield  {journal}
  {\bibinfo  {journal} {J. Opt. Soc. Am. B}\ }\textbf {\bibinfo {volume} {2}},\
  \bibinfo {pages} {552} (\bibinfo {year} {1985})}\BibitemShut {NoStop}%
\bibitem [{\citenamefont {Holling}(1959{\natexlab{a}})}]{Holling:1959a}%
  \BibitemOpen
  \bibfield  {author} {\bibinfo {author} {\bibfnamefont {C.~S.}\ \bibnamefont
  {Holling}},\ }\bibfield  {title} {\bibinfo {title} {The components of
  predation as revealed by a study of small-mammal predation of the european
  pine sawfly},\ }\href@noop {} {\bibfield  {journal} {\bibinfo  {journal}
  {Canad. Entomol.}\ }\textbf {\bibinfo {volume} {91}},\ \bibinfo {pages}
  {293–320} (\bibinfo {year} {1959}{\natexlab{a}})}\BibitemShut {NoStop}%
\bibitem [{\citenamefont {Holling}(1959{\natexlab{b}})}]{Holling:1959b}%
  \BibitemOpen
  \bibfield  {author} {\bibinfo {author} {\bibfnamefont {C.~S.}\ \bibnamefont
  {Holling}},\ }\bibfield  {title} {\bibinfo {title} {Some characteristics of
  simple types of predation and parasitism},\ }\href@noop {} {\bibfield
  {journal} {\bibinfo  {journal} {Canad. Entomol.}\ }\textbf {\bibinfo {volume}
  {91}},\ \bibinfo {pages} {385} (\bibinfo {year}
  {1959}{\natexlab{b}})}\BibitemShut {NoStop}%
\bibitem [{\citenamefont {Jiang}\ \emph {et~al.}(2018)\citenamefont {Jiang},
  \citenamefont {Huang}, \citenamefont {Seager}, \citenamefont {Lin},
  \citenamefont {Grebogi}, \citenamefont {Hastings},\ and\ \citenamefont
  {Lai}}]{JHSLGHL:2018}%
  \BibitemOpen
  \bibfield  {author} {\bibinfo {author} {\bibfnamefont {J.}~\bibnamefont
  {Jiang}}, \bibinfo {author} {\bibfnamefont {Z.-G.}\ \bibnamefont {Huang}},
  \bibinfo {author} {\bibfnamefont {T.~P.}\ \bibnamefont {Seager}}, \bibinfo
  {author} {\bibfnamefont {W.}~\bibnamefont {Lin}}, \bibinfo {author}
  {\bibfnamefont {C.}~\bibnamefont {Grebogi}}, \bibinfo {author} {\bibfnamefont
  {A.}~\bibnamefont {Hastings}},\ and\ \bibinfo {author} {\bibfnamefont
  {Y.-C.}\ \bibnamefont {Lai}},\ }\bibfield  {title} {\bibinfo {title}
  {Predicting tipping points in mutualistic networks through dimension
  reduction},\ }\href@noop {} {\bibfield  {journal} {\bibinfo  {journal} {Proc.
  Natl. Acad. Sci. (USA)}\ }\textbf {\bibinfo {volume} {115}},\ \bibinfo
  {pages} {E639} (\bibinfo {year} {2018})}\BibitemShut {NoStop}%
\bibitem [{\citenamefont {Liu}\ \emph {et~al.}(2024)\citenamefont {Liu},
  \citenamefont {Wang}, \citenamefont {Vaidya}, \citenamefont {Ruehle},
  \citenamefont {Halverson}, \citenamefont {Soljači\'{c}}, \citenamefont
  {Hou},\ and\ \citenamefont {Tegmark}}]{LWVRHSHT:2024}%
  \BibitemOpen
  \bibfield  {author} {\bibinfo {author} {\bibfnamefont {Z.}~\bibnamefont
  {Liu}}, \bibinfo {author} {\bibfnamefont {Y.}~\bibnamefont {Wang}}, \bibinfo
  {author} {\bibfnamefont {S.}~\bibnamefont {Vaidya}}, \bibinfo {author}
  {\bibfnamefont {F.}~\bibnamefont {Ruehle}}, \bibinfo {author} {\bibfnamefont
  {J.}~\bibnamefont {Halverson}}, \bibinfo {author} {\bibfnamefont
  {M.}~\bibnamefont {Soljači\'{c}}}, \bibinfo {author} {\bibfnamefont {T.~Y.}\
  \bibnamefont {Hou}},\ and\ \bibinfo {author} {\bibfnamefont {M.}~\bibnamefont
  {Tegmark}},\ }\bibfield  {title} {\bibinfo {title} {{KAN: Kolmogorov-Arnold}
  networks},\ }\href@noop {} {\bibfield  {journal} {\bibinfo  {journal}
  {arXiv:2404.19756}\ } (\bibinfo {year} {2024})}\BibitemShut {NoStop}%
\bibitem [{\citenamefont {Kolmogorov}(1957)}]{Kolmogorov:1957}%
  \BibitemOpen
  \bibfield  {author} {\bibinfo {author} {\bibfnamefont {A.~N.}\ \bibnamefont
  {Kolmogorov}},\ }\bibfield  {title} {\bibinfo {title} {On the representation
  of continuous functions of many variables by superposition of continuous
  functions of one variable and addition},\ }\href@noop {} {\bibfield
  {journal} {\bibinfo  {journal} {Rus. Acad. Sci.}\ }\textbf {\bibinfo {volume}
  {114}},\ \bibinfo {pages} {953} (\bibinfo {year} {1957})}\BibitemShut
  {NoStop}%
\bibitem [{\citenamefont {Givental}\ \emph {et~al.}(2009)\citenamefont
  {Givental}, \citenamefont {Khesin}, \citenamefont {Marsden}, \citenamefont
  {Varchenko}, \citenamefont {Vassiliev}, \citenamefont {Viro},\ and\
  \citenamefont {Zakalyukin}}]{Arnold:2009}%
  \BibitemOpen
  \bibinfo {editor} {\bibfnamefont {A.~B.}\ \bibnamefont {Givental}}, \bibinfo
  {editor} {\bibfnamefont {B.~A.}\ \bibnamefont {Khesin}}, \bibinfo {editor}
  {\bibfnamefont {J.~E.}\ \bibnamefont {Marsden}}, \bibinfo {editor}
  {\bibfnamefont {A.~N.}\ \bibnamefont {Varchenko}}, \bibinfo {editor}
  {\bibfnamefont {V.~A.}\ \bibnamefont {Vassiliev}}, \bibinfo {editor}
  {\bibfnamefont {O.~Y.}\ \bibnamefont {Viro}},\ and\ \bibinfo {editor}
  {\bibfnamefont {V.~M.}\ \bibnamefont {Zakalyukin}},\ eds.,\ \bibinfo {title}
  {On the representation of functions of several variables as a superposition
  of functions of a smaller number of variables},\ in\ \href
  {https://doi.org/10.1007/978-3-642-01742-1_5} {\emph {\bibinfo {booktitle}
  {Collected Works: Representations of Functions, Celestial Mechanics and KAM
  Theory, 1957--1965}}}\ (\bibinfo  {publisher} {Springer Berlin Heidelberg},\
  \bibinfo {address} {Berlin, Heidelberg},\ \bibinfo {year} {2009})\ pp.\
  \bibinfo {pages} {25--46}\BibitemShut {NoStop}%
\bibitem [{\citenamefont {Braun}\ and\ \citenamefont
  {Griebel}(2009)}]{BG:2009}%
  \BibitemOpen
  \bibfield  {author} {\bibinfo {author} {\bibfnamefont {J.}~\bibnamefont
  {Braun}}\ and\ \bibinfo {author} {\bibfnamefont {M.}~\bibnamefont
  {Griebel}},\ }\bibfield  {title} {\bibinfo {title} {On a constructive proof
  of kolmogorov’s superposition theorem},\ }\href@noop {} {\bibfield
  {journal} {\bibinfo  {journal} {Const. Appro.}\ }\textbf {\bibinfo {volume}
  {30}},\ \bibinfo {pages} {653} (\bibinfo {year} {2009})}\BibitemShut
  {NoStop}%
\bibitem [{SI()}]{SI}%
  \BibitemOpen
  \href@noop {} {\bibinfo  {journal} {Supplementary Information contains a
  detailed description of the KAN framework for model discovery of nonlinear
  dynamical systems, statistical analysis of the KAN generated attractors, and
  additional examples}\ }\BibitemShut {NoStop}%
\bibitem [{\citenamefont {McCann}\ and\ \citenamefont
  {Yodzis}(1994)}]{MCcann1994}%
  \BibitemOpen
\bibfield  {journal} {  }\bibfield  {author} {\bibinfo {author} {\bibfnamefont
  {K.}~\bibnamefont {McCann}}\ and\ \bibinfo {author} {\bibfnamefont
  {P.}~\bibnamefont {Yodzis}},\ }\bibfield  {title} {\bibinfo {title}
  {Nonlinear dynamics and population disappearances},\ }\href@noop {}
  {\bibfield  {journal} {\bibinfo  {journal} {Am. Nat}\ }\textbf {\bibinfo
  {volume} {144}},\ \bibinfo {pages} {873} (\bibinfo {year}
  {1994})}\BibitemShut {NoStop}%
\bibitem [{\citenamefont {Binmore}(1982)}]{binmore1982mathematical}%
  \BibitemOpen
  \bibfield  {author} {\bibinfo {author} {\bibfnamefont {K.~G.}\ \bibnamefont
  {Binmore}},\ }\href@noop {} {\emph {\bibinfo {title} {Mathematical Analysis:
  a straightforward approach}}}\ (\bibinfo  {publisher} {Cambridge University
  Press},\ \bibinfo {year} {1982})\BibitemShut {NoStop}%
\end{thebibliography}

%
\end{document}